\documentclass[accepted]{uai2025} 
\pdfoutput=1
                        
\usepackage[american]{babel}

\usepackage[sort&compress]{natbib}

\usepackage{mathtools} 
\usepackage{booktabs} 
\usepackage{tikz} 
\usepackage{amsmath,amssymb,amsfonts}
\usepackage{xifthen}
\usepackage{subcaption}
\usepackage{graphicx}
\usepackage{textcomp}
\usepackage{yhmath}
\usepackage[ruled,vlined]{algorithm2e}
\usepackage{bm}
\usepackage{cancel}
\usepackage{xcolor}
\usepackage{textcase}
\usepackage{siunitx}
\usepackage{array}
\usepackage{multirow}
\usepackage{gensymb}
\usepackage{tabularx}
\usepackage{extarrows}
\usepackage{booktabs}
\usepackage{colortbl}
\usepackage{pgfplots}
\usepackage{pgfplotstable}

\pgfplotsset{compat=newest}
\usetikzlibrary{fadings}
\usetikzlibrary{bayesnet}
\usetikzlibrary{patterns}
\usetikzlibrary{shadows.blur}
\usetikzlibrary{calc}
\usetikzlibrary{shapes, 3d, backgrounds, fit, arrows.meta, positioning, shapes.geometric}
\usepackage{custom_commands}

\allowdisplaybreaks

\title{Partially Observable Gaussian Process Network and Doubly Stochastic Variational Inference}

\author[1]{\href{mailto:<saksham.kiroriwal@iosb.fraunhofer.de>?Subject=Your paper: Partially Observable Gaussian Process Network and Doubly Stochastic Variational Inference}{Saksham Kiroirwal}}
\author[1]{\href{mailto:<julius.pfrommer@iosb.fraunhofer.de>?Subject=Your paper: Partially Observable Gaussian Process Network and Doubly Stochastic Variational Inference}{Julius Pfrommer}}
\author[1,2]{\href{mailto:<juergen.beyerer@iosb.fraunhofer.de>?Subject=Your paper: Partially Observable Gaussian Process Network and Doubly Stochastic Variational Inference}{Jürgen Beyerer}}
\affil[1]{%
    Cognitive Industrial Systems\\
    Fraunhofer IOSB\\
    Karlsruhe, Germany
}
\affil[2]{%
    Karlsruhe Institute of Technology\\
    Karlsruhe, Germany
}

\begin{document}
\maketitle

\begin{abstract}
    To reduce the curse of dimensionality for Gaussian processes (GP), they can be decomposed into a Gaussian Process Network (GPN) of coupled subprocesses with lower dimensionality. In some cases, intermediate observations are available within the GPN. However, intermediate observations are often indirect, noisy, and incomplete in most real-world systems. This work introduces the Partially Observable Gaussian Process Network (POGPN) to model real-world process networks. We model a joint distribution of latent functions of subprocesses and make inferences using observations from all subprocesses. POGPN incorporates observation lenses (observation likelihoods) into the well-established inference method of deep Gaussian processes. We also introduce two training methods for POPGN to make inferences on the whole network using node observations. The application to benchmark problems demonstrates how incorporating partial observations during training and inference can improve the predictive performance of the overall network, offering a promising outlook for its practical application.
\end{abstract}

\section{Introduction}\label{sec:intro}
Conventionally, a process is considered a single-process black box with input(s) and some output(s) being observed. However, systems are hardly single-process and comprise multiple sub-processes where intermediate outputs from each subprocess can be observed~\cite {Fenner2005Optimal}. The respective sub-processes can be stochastic as well. Figure~\ref{fig:multi_process} shows an example two-process system with root inputs $\cusvector{\params}^{(1)}$, observed intermediate output(s) $\cusvector{\tilde{\obsouts}}^{(1)}$, and observed final output(s) $\cusvector{\tilde{\obsouts}}^{(2)}$.
\begin{figure}[ht]
    \centering
    \tikzset{every picture/.style={line width=0.7pt}} 
\tikzset{plate caption/.append style={left=2pt of #1.south east, yshift=0.2cm}}
\begin{tikzpicture}[
    block/.style={rectangle, draw, rounded corners, minimum height=3.2em, minimum width=4em, thick},
    normal_arrow/.style={-Stealth},
    obs/.style={circle, draw, minimum size=6.5mm, align=center, inner sep=0.7pt}
    every node/.style={font=\normalsize},
    every picture/.style={line width=1pt}
    ]

    \node[block, align=center] (process1) {Subprocess \\ $\process^{(1)}$};
    \node[block, align=center, right=of process1, xshift=-0.35cm] (process2) {Subprocess \\ $\process^{(2)}$};
    \node[left=of process1, xshift=0.4cm] (input1) {$\cusvector{\params}^{(1)}$};
    \node[below=of process2, yshift=0.1cm] (input2) {$\cusvector{\params}^{(2)}$};
    \node[below=of process1, xshift=-1cm, yshift=0.8cm] (noise1) {$\cusvector{\tilde{\noise}}^{(1)}$};
    \node[below=of process2, xshift=-1cm, yshift=0.8cm] (noise2) {$\cusvector{\tilde{\noise}}^{(2)}$};
    \node[right=of process2, xshift=0.15cm] (f2_psuedo) {};
    \node[draw=none, right=of f2_psuedo, xshift=-0.8cm] (y2) {$\cusvector{\tilde{\obsouts}}^{(2)}$};
    \draw[draw=darkgreen, normal_arrow] (input1) -- (process1);
    \draw[draw=darkgreen, normal_arrow] (input2) -- (process2);
    \draw[draw=red, normal_arrow] (process1) -- node[above] (f1){$\cusvector{\trueouts}^{(1)}$} (process2);
    \node[draw=none, above=of f1, yshift=-0.7cm] (y1) {$\cusvector{\tilde{\obsouts}}^{(1)}$};
    \draw[draw=black, normal_arrow] (noise1) -- (process1);
    \draw[draw=black, normal_arrow] (noise2) -- (process2);
    \draw[draw=red, -] (process2) -- node[above] (f2) {$\cusvector{\trueouts}^{(2)}$}(f2_psuedo.center); 
    \draw[draw=blue, normal_arrow] (f2_psuedo.center) -- (y2); 
    \draw[draw=blue, normal_arrow] (f1) -- (y1);
    \plate [dashed, inner sep=-0.02cm] {process} {(process1) (process2) (y1) (f2_psuedo.center) (noise1) (noise2)} {Process: $\Process$};

\end{tikzpicture}
    \vspace{-1em}
    \caption{Example process network where stochastic subprocesses are coupled by the latent state $\cusvector{\processfunction}^{(\cdot)}$ which is partially observable as $\cusvector{\obsouts}^{(\cdot)}$. Along with the input from the parent, a subprocess can also have adjustable input $\cusvector{\params}^{(\cdot)}.$}
    \vspace{-1em}
    \label{fig:multi_process}
\end{figure}
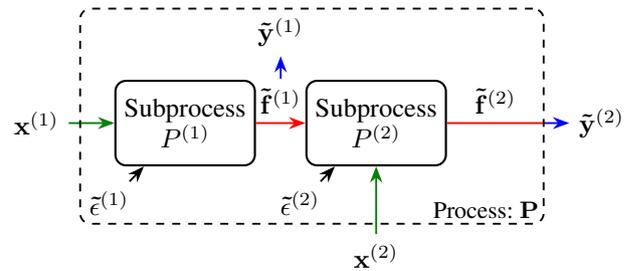
\begin{figure}[!ht]
    \centering
    \begin{subfigure}{\columnwidth}
        \centering
        \tikzset{every picture/.style={line width=0.7pt}} 
\begin{tikzpicture}[node distance=10mm and 10mm,
        process/.style={circle, draw, minimum size=6.5mm, align=center, inner sep=0.7pt},
        every node/.style={font=\normalsize},
        arr/.style={-Stealth},
        obs/.style={circle, draw, fill=gray!30, minimum size=6.5mm, align=center, inner sep=0.7pt}
    ]
    \node[obs] (input1) {$\cusvector{\params}^{(1)}$};
    \node[process] (f1) [right=of input1, xshift=-0.2cm] {$\cusvector{\processfunction}^{(1)}$};
    \node[obs] (y1) [right=of f1] {$\cusvector{\tilde{\obsouts}}^{(1)}$};
    \node[process] (f2) [right=of y1, xshift=0.2cm] {$\cusvector{\processfunction}^{(2)}$};
    \node[obs] (y2) [right=of f2, xshift=0.2cm] {$\cusvector{\tilde{\obsouts}}^{(2)}$};
    \node[obs] (input2) [below=of f2, yshift=0.3cm] {$\cusvector{\params}^{(2)}$};

    \draw[arr, draw=darkgreen] (input2) -- (f2);
    \draw[arr, draw=darkgreen] (input1) -- node[above, yshift=0.35cm]
    {$\probability(\cusvector{\processfunction}^{(1)}\vert \cusvector{\params}^{(1)})$} (f1);
    \draw[arr, draw=blue] (f1) -- node[below, yshift=-0.35cm]
    {$\probability(\cusvector{\obsouts}^{(1)}\vert\cusvector{\processfunction}^{(1)})$} (y1);
    \draw[arr, draw=red] (y1) -- node[above, yshift=0.35cm]
    {$\probability(\cusvector{\processfunction}^{(2)}\vert\cusvector{\params}^{(2)}, \cusvector{\tilde{\obsouts}}^{(1)})$} (f2);
    \draw[arr, draw=blue] (f2) -- node[below, yshift=-0.35cm, xshift=0.2cm]
    {$\probability(\cusvector{\obsouts}^{(2)}\vert\cusvector{\processfunction}^{(2)})$} (y2);

\end{tikzpicture}
        \caption{Existing Gaussian Process Network (GPN)}
        \label{fig:toy_process_gpn}
    \end{subfigure}
    \begin{subfigure}{\columnwidth}
        \centering
        \tikzset{every picture/.style={line width=0.7pt}} 
\begin{tikzpicture}[node distance=10mm and 10mm,
        process/.style={circle, draw, minimum size=6.5mm, align=center, inner sep=0.7pt},
        every node/.style={font=\normalsize},
        arr/.style={-Stealth},
        obs/.style={circle, draw, fill=gray!30, minimum size=6.5mm, align=center, inner sep=0.7pt}
    ]
    \node[obs] (input1){$\cusvector{\params}^{(1)}$};
    \node[process] (f1) [right=of input1, xshift=-0.2cm] {$\cusvector{\processfunction}^{(1)}$};
    \node[obs] (y1) [right=of f1, yshift=1.5cm] {$\cusvector{\tilde{\obsouts}}^{(1)}$};
    \node[process, draw=none] (midpoint_f1) [right=of f1, xshift=-0.33cm] {};
    \node[process] (f2) [right=of midpoint_f1, xshift=0.5cm] {$\cusvector{\processfunction}^{(2)}$};
    \node[obs] (input2) [right=of f2, xshift=-0.1cm] {$\cusvector{\params}^{(2)}$};
    \node[process, draw=none] (midpoint_y1) [right=of y1, xshift=0cm] {};
    \node[obs] (y2) [right=of midpoint_y1] {$\cusvector{\tilde{\obsouts}}^{(2)}$};

    \draw[arr, draw=darkgreen] (input1) -- node[below, yshift=-0.25cm, xshift=-0.25cm]
    {$\probability(\cusvector{\processfunction}^{(1)}\vert\cusvector{\params}^{(1)})$} (f1);
    \draw[arr, draw=darkgreen] (input2) -- (f2);
    \draw[arr, draw=blue] (f1) -- node[above, sloped, yshift=0.25cm]
    {$\probability(\cusvector{\obsouts}^{(1)}\vert\cusvector{\processfunction}^{(1)})$} (y1);
    \draw[arr, draw=red] (f1) -- node[below, yshift=-0.25cm, xshift=0.25cm]
    {$\probability(\cusvector{\processfunction}^{(2)}\vert\cusvector{\params}^{(2)}, \cusvector{\processfunction}^{(1)})$} (f2);
    \draw[arr, draw=blue] (f2) -- node[above, sloped, yshift=0.25cm]
    {$\probability(\cusvector{\obsouts}^{(2)}\vert\cusvector{\processfunction}^{(2)})$} (y2);

\end{tikzpicture}
        \caption{Partially Observable Gaussian Process Network (POGPN)}
        \label{fig:toy_process_pogpn}
    \end{subfigure}
    \caption{Comparison of GP network and POGPN. Gray nodes represent observed outputs (likelihood), and white nodes represent latent outputs (GP).}
    \vspace{-1em}
    \label{fig:gpn_pogpn_comaprison}
\end{figure}
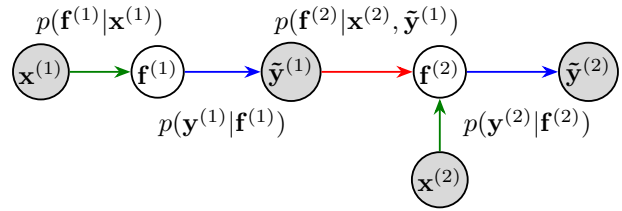
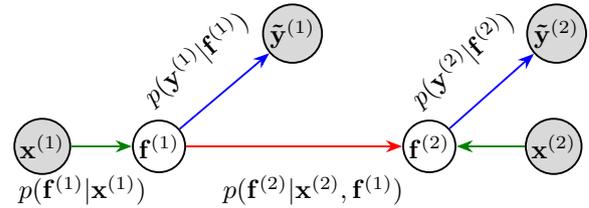

Gaussian processes (GP) are a popular probabilistic framework to model nonlinear input-output dependencies. They have been widely used in various applications, such as the monitoring of key performance indicators for processes~\citep{Kontar2017Estimation}, control~\citep{Likar2007Predictive} and Bayesian optimization~\citep{Frazier2015Bayesian}. Work by~\cite{astudillo2021bayesian, aglietti2020causal, sussex2022model, kusakawa2022bayesian, aglietti2020multi, SakshamJPSS} show improved results when intermediate observations are included in modeling and Bayesian optimization. We call the generic class of models used for these results the Gaussian Process Network (GPN).

The GPNs employ a GP network as a Directed Acyclic Graph (DAG), where each node represents a sub-process for which an intermediate observation is recorded. The node GPs are trained to predict the respective node observation(s). The implementation considers GPs to be conditionally independent given the respective input-output pairs using closed-form Marginal Log Likelihood (MLL). However, the existing GPNs have some limitations. In most physical processes, we can observe the output-space only partially with an observation noise. For this reason, the existing GPN training methods can only be used in rare situations where we can make noise-free observations. Also, the models do not scale well to the case of non-Gaussian likelihood observations or large datasets. These limitations underscore the need for a new model.

Section~\ref{sec:multi_process} explains the definition of a DAG process with partial observability, which leads to section~\ref{sec:gp_related_work}, where the existing GPN work and its limitations are discussed. Section~\ref{sec:pogpn} and~\ref{sec:experiments} discusses and shows experiment results of our contributions :
\begin{itemize}
    \item Partially Observable Gaussian Process Network (POPGN): a real-world process-inspired model that overcomes the limitations of the current GPNs.
    \item Doubly stochastic variational inference for POGPN using Evidence Lower BOund (ELBO).
    \item Inference using Predictive Log Likelihood (PLL).
    \item Training methods to condition POGPN on partial observations of nodes/subprocesses.
\end{itemize}

\section{Process Network with Partial Observability}\label{sec:multi_process}
We consider a stochastic process $\Process$ comprised of subprocesses, $\{\process^{(\numprocess)}\}$ for $\numprocess\in\numprocessset$, (which can be stochastic) as shown in Figure~\ref{fig:multi_process}. The process of stochasticity can come from either a lack of knowledge of the process or other hidden influences or random noise $\cusvector{\tilde{\noise}}^{(\numprocess)}\sim\probability(\cusvector{\noise}^{(\numprocess)})$. Process $\Process$ is a DAG where the nodes represent the subprocesses. Each subprocess $\process^{(\numprocess)}$ is governed by a transformation function $\cusvector{\processfunction}^{(\numprocess)}$ which takes as input(s) some adjustable parameters $\cusvector{\params}^{(\numprocess)}$ (represented with green arrow in Figure~\ref{fig:multi_process}) and the output(s) of parent subprocesses $\process^{\nodeparent{\numprocess}}$, where $\nodeparent{\numprocess}$ represents the direct parents of $\numprocess$. Using the transformation function, the contactenated input(s) $(\cusvector{\params}^{(\numprocess)}, \process^{\nodeparent{\numprocess}})$ are transformed into output(s) $\cusvector{\trueouts}^{(\numprocess)}$ (represented with red arrow).

Figure~\ref{fig:multi_process} shows an example stochastic process with two subprocesses that can be expressed using a distribution over the function space, which the transformation function $\cusvector{\processfunction}^{(\numprocess)}$ is a sampled from. Using this definition, the outputs of the two-process system shown in Figure~\ref{fig:multi_process} as red arrows can be redefined as
\begin{equation*}
      \cusvector{\trueouts}^{(\numprocess)} \sim \probability\big(\cusvector{\processfunction}^{(\numprocess)}\vert\cusvector{\params}^{(\numprocess)}, \cusvector{\processfunction}^{(\nodeparent{\numprocess})}\big), \forall \numprocess\in\{1, 2\}.
\end{equation*}
In most cases, the output(s) $\cusvector{\trueouts}^{(\numprocess)}$ of a subprocess $\process^{(\numprocess)}$ cannot be fully observed and are observed indirectly/partially using an "observation lens" (represented with blue arrow) as $\cusvector{\tilde{\obsouts}}^{(\numprocess)}$, hence the name partially observable process network. An example of such an observation lens can be the Gaussian observation noise of a sensor. In probabilistic modeling, the "observation lens" is often modeled as the likelihood function $\cusvector{\tilde{\obsouts}}^{(\numprocess)}\sim\probability(\cusvector{\obsouts}^{(\numprocess)}\vert\cusvector{\processfunction}^{(\numprocess)})$, which could be as simple as additive Gaussian noise or something more complex. The true output(s) $\cusvector{\trueouts}^{(\numprocess)}$ remain latent. We assume that, the observation lens is always present whenever an output is referred to as "observed" unless stated as "latent" or "true" output. The "latent" output of a parent subprocess becomes the input for a child subprocess.

At this point we are able to define each subprocess $\process^{(\numprocess)}$ using a tuple $\langle\probability\big(\cusvector{\processfunction}^{(\numprocess)}\vert\cusvector{\params}^{(\numprocess)}, \cusvector{\trueouts}^{\nodeparent{\numprocess}}\big), \probability(\cusvector{\obsouts}^{(\numprocess)}\vert\cusvector{\processfunction}^{(\numprocess)})\rangle$, where $\cusvector{\params}^{(\numprocess)}$ represents the adjustable input parameters, $\cusvector{\trueouts}^{\nodeparent{\numprocess}}$ represents the latent output(s) from the parent subprocess(es), $\probability\big(\cusvector{\processfunction}^{(\numprocess)}\vert\cusvector{\params}^{(\numprocess)}, \cusvector{\trueouts}^{\nodeparent{\numprocess}}\big)$ represents the probability distribution over the transformation function and $\probability(\cusvector{\obsouts}^{(\numprocess)}\vert\cusvector{\processfunction}^{(\numprocess)})$ represents the observation likelihood or lens with which the latent output of the subprocess $\process^{(\numprocess)}$ is observed.

The process network $\Process$ can be represented using a DAG, $\graph$. The nodes are topologically ordered such that $\numprocess'<\numprocess ,\forall \numprocess' \in \nodeparent{\numprocess}$ for all $\numprocess\in\numprocessset$. We use the terms subprocess and node interchangeably to represent a subprocess in $\graph$. $\cusvector{\params}^{(\numprocess)}$ are addressed as adjustable input nodes or parameters. The final/end process of the process $\Process$ is defined as the subprocess(es) $\process^{(\numprocess)}$ for which there are no child nodes. The corresponding observed output(s) $\cusvector{\tilde{\obsouts}}^{(\numprocess)}$ are called the observed final output(s).

The problem statement is to model the subprocess output(s), and the end process output using the adjustable inputs of different subprocess(es) and all indirect observations of the process network made using different observation likelihoods. We assume that the data generation DAG or the causal path is known. We refrain from augmenting the intermediate observations with adjustable inputs to avoid blowing up the input dimensionality for the used model. We discuss the proposed solution in section~\ref{sec:pogpn}.

\section{Background}\label{sec:background}
In this section we discuss Gaussian processes (GP), deep GP, and existing GP networks as well as their limitations.
\subsection{Gaussian process}\label{sec:stgp}
Considering the multi-process system in Figure~\ref{fig:multi_process}, it is conventionally modeled as a single-process black box with inputs $\cusvector{\params}_{\numobservation}:=\cusvector{\params}_{\numobservation}^{(1)} \in \paramsset$ and observed outputs $\tilde{\obsouts}_{\numobservation}=\tilde{\obsouts}_{\numobservation}^{(2)}$. A Gaussian process represents a stochastic process as a distribution over the infinite-dimensional functions, which can be evaluated at an input location in $\paramsset$. A finite set of observed outputs evaluated at respective input locations represents a multivariate normal distribution. For a given input location, $\cusvector{\params}_{\numobservation}$, of dimension $\dims{\params}$, the GP prior is represented as
\begin{equation}\label{eq:exact_gp_prior}
      \underbrace{\probability(\processfunction_{\numobservation}\vert\cusvector{\params}_{\numobservation})}_{\text{exact GP prior}} =\mathcal{N}\big(\gpmean(\cusvector{\params}_{\numobservation}), \gpkernel(\cusvector{\params}_{\numobservation}, \cusvector{\params}_{\numobservation}')\big),
\end{equation}
where $\gpmean(\cdot):\mathbb{R}^{\dims{\params}}\rightarrow\mathbb{R}$ is the mean function and $\gpkernel(\cdot,\cdot'):\paramsset\times\paramsset\rightarrow\mathbb{R}$ is the covariance function or kernel~\citep{rasmussen2003gaussian}.

For $\Numobservation$ observations, the hyperparameters of the GP are optimized by minimizing the negative MLL for the observed outputs using~\eqref{eq:exact_gp_prior} as
\begin{equation}\label{eq:STGP_MLL_exact}
      \mathcal{L}_\text{GP} = - \sum^{\Numobservation}_{\numobservation=1}\log\mathbb{E}_{\probability(\processfunction_{\numobservation}\vert\cusvector{\params}_{\numobservation})}\big[\probability(\obsouts_{\numobservation}\vert\processfunction_{\numobservation})\big],
\end{equation}
which can be calculated in closed form for a Gaussian likelihood and scales with $\mathcal{O}(\Numobservation^3)$~\citep{rasmussen2003gaussian}. The concepts of the single task GP can be extended to vector-valued stochastic process $\cusvector{\processfunction}_{\numobservation}(\cdot):\mathbb{R}^{\dims{\params}}\rightarrow\mathbb{R}^{\dims{\obsouts}}$ in which the observed output is a vector, of dimension $\dims{\obsouts}$, for each input location. A popular way of modeling correlation between the outputs is using the Linear Model of Coregionalization (LMC) as explained by~\cite{alvarez2012kernels,van2020framework}.

\subsection{Stochastic Variational Gaussian Process}\label{sec:SVGP}

Stochastic Variational Gaussian Processes (SVGP) by~\cite{hensman2013gaussian, hensman2015scalable} are useful for large datasets and non-Gaussian likelihoods. It assumes a set of inducing locations, $\cusmatrix{\Inducinglocations}=\{\cusvector{\inducinglocations}_{\numinducing}\}^{\Numinducing}_{\numinducing=1}\in \paramsset$, and
inducing points which are, function value evaluations $\cusvector{\inducingpoints}=\{\inducingpoints_{\numinducing}\}_{\numinducing=1}^{\Numinducing}$ at $\{\cusvector{\inducinglocations}_{\numinducing}\}_{\numinducing=1}^{\Numinducing}$. The joint distribution of inducing points $\cusvector{\inducingpoints}$ can be represented as
\begin{equation}\label{eq:svgp_inducing_prior}
      \probability(\cusvector{\inducingpoints};\cusmatrix{\Inducinglocations}) = \mathcal{N}\big(\gpmean(\cusmatrix{\Inducinglocations}), \gpkernel(\cusmatrix{\Inducinglocations}, \cusmatrix{\Inducinglocations}')\big).
\end{equation}

Additionally, the inducing points are also assumed to have a marginal distribution $\variationalprob(\cusvector{\inducingpoints})=\mathcal{N}\big(\cusvector[bm]{\distmean}_{\cusvector{\inducingpoints}}, \cusmatrix[bm]{\distcov}_{\cusvector{\inducingpoints}}\big)$.
A joint multivariate normal distribution $\probability(\processfunction_{\numobservation},\cusvector{\inducingpoints}\vert\cusvector{\params}_{\numobservation},\cusmatrix{\Inducinglocations})$ can be defined using~\eqref{eq:exact_gp_prior} and~\eqref{eq:svgp_inducing_prior} . For further details, readers are encouraged to refer to~\cite{leibfried2020tutorial}.

SVGP approximates the exact posterior with a variational posterior \begin{align}\label{eq:svgp_marginal_qf}
      \variationalprob(\processfunction_{\numobservation}\vert \cusvector{\params}_{\numobservation}, \cusmatrix{\Inducinglocations}) & =\mathbb{E}_{\variationalprob(\cusvector{\inducingpoints})}\big[\probability(\processfunction_{\numobservation}\vert\cusvector{\inducingpoints})\big]=\mathcal{N}\big(\distmean_{\variationalprob(\processfunction)}, \distcov_{\variationalprob(\processfunction)}\big),
\end{align}
where
\begin{align}\label{eq:svgp_marginal_qf_dist}
      \distmean_{\variationalprob(\cusvector{\processfunction})}            & = \gpmean(\cusvector{\params}_{\numobservation}) + \cusmatrix[bm]{\SvgpDistConst}(\cusvector{\params}_{\numobservation})^{\top}\big(\cusvector[bm]{\distmean}_{\cusvector{\inducingpoints}}-\gpmean(\cusmatrix{\Inducinglocations})\big),\nonumber                                                                                                                                                \\
      \distcov_{\variationalprob(\cusvector{\processfunction})}             & = \gpkernel(\cusvector{\params}_{\numobservation}, \cusvector{\params}_{\numobservation}) - \cusmatrix[bm]{\SvgpDistConst}(\cusvector{\params}_{\numobservation})^{\top}\big(\gpkernel(\cusmatrix{\Inducinglocations}, \cusmatrix{\Inducinglocations})-\cusmatrix[bm]{\distcov}_{\cusvector{\inducingpoints}}\big)\cusmatrix[bm]{\SvgpDistConst}(\cusvector{\params}_{\numobservation}),\nonumber \\
      \cusmatrix[bm]{\SvgpDistConst}(\cusvector{\params}_{\numobservation}) & =\gpkernel(\cusmatrix{\Inducinglocations}, \cusmatrix{\Inducinglocations})^{-1}\gpkernel(\cusmatrix{\Inducinglocations}, \cusvector{\params}_{\numobservation}).
\end{align}
Since the variational posterior $\variationalprob(\processfunction_{\numobservation}; \cusvector{\params}_{\numobservation}, \cusmatrix{\Inducinglocations})$, abbreviated as $\variationalprob(\processfunction_{\numobservation})$,
is conditioned on a fixed number of inducing points, SVGP can scale and be used for non-Gaussian likelihoods.
Hyperparameters are optimized by minimizing the negative Evidence Lower BOund (ELBO)
\begin{equation}\label{eq:svgp_elbo}
      \mathcal{L}_{\substack{\text{SVGP} \\ \text{ELBO}}} = -\sum^{\Numobservation}_{\numobservation=1} \Big[\mathbb{E}_{\variationalprob(\processfunction_{\numobservation})}[\log \probability(\obsouts_{\numobservation}\vert\processfunction_{\numobservation})]\Big]+\klconst \text{KL}\big(\variationalprob(\cusvector{\inducingpoints})\vert\vert\probability(\cusvector{\inducingpoints})\big),
\end{equation}
where~\eqref{eq:svgp_marginal_qf} and~\eqref{eq:svgp_marginal_qf_dist} are used to calculate $\mathbb{E}_{\variationalprob(\processfunction_{\numobservation})}$~\citep{hensman2015scalable}. ~\cite{jankowiak2020parametric} introduced another variational loss, namely, Parametric Predictive GP Regressor (PPGPR), which minimizes the negative Predictive Log Likelihood (PLL) for SVGP as
\begin{equation}\label{svgp:pll}
      \mathcal{L}_{\substack{\text{SVGP} \\ \text{PLL}}} = -\sum^{\Numobservation}_{\numobservation=1} \Big[\log \mathbb{E}_{\variationalprob(\processfunction_{\numobservation})}[\probability(\obsouts_{\numobservation}\vert\processfunction_{\numobservation})]\Big]+\klconst \text{KL}\big(\variationalprob(\cusvector{\inducingpoints})\vert\vert\probability(\cusvector{\inducingpoints})\big).
\end{equation}
The PPGPR claims to provide better predictive performance than ELBO. However, for non-conjugate likelihoods, the expectation for PLL cannot be calculated in closed form~\citep{jankowiak2020deep}.~\cite{van2020framework} show how inducing points approximation can be extended to multi-task GP.

\subsection{Deep Gaussian Process}\label{sec:DGP}
Deep Gaussian Process (DGP) introduced by~\cite{damianou2013deep} provides a hierarchical model, in which independent GPs are stacked in layers as $\{\cusvector{\processfunction}^{(\numDGPlayer)}\in\mathbb{R}^{\dims{\numDGPlayer}}\}^{\NumDGPlayer}_{\numDGPlayer=1}$. Vector-valued outputs $\cusvector{\processfunction}_{\numobservation}^{(\numDGPlayer-1)}$ from the previous layer GPs become the inputs for the next layer output $\cusvector{\processfunction}_{\numobservation}^{(\numDGPlayer)}$.
A prominent inference method for DGP is doubly stochastic variational inference by~\cite{salimbeni2017doubly}, which builds on the inference method of SVGP.

Similar to the idea of inducing points in the input domain of SVGP, inducing points $\{\cusmatrix{\Inducingpoints}^{(\numDGPlayer)}\in\mathbb{R}^{\Numinducing(\numDGPlayer)\times\dims{\numDGPlayer}}\}^{\NumDGPlayer}_{\numDGPlayer=1}$ are introduced at inducing locations $\{\cusmatrix{\Inducinglocations}^{(\numDGPlayer-1)}\in\mathbb{R}^{\Numinducing(\numDGPlayer)\times\dims{\numDGPlayer-1}}\}^{\NumDGPlayer}_{\numDGPlayer=1}$, where $\dims{0}=\dims{\params}$ and $\Numinducing(\numDGPlayer)$ is the number of inducing points for layer $\numDGPlayer$. Since the GPs are independent of each other~\cite{salimbeni2017doubly}, the marginal distribution for a layer depends only on the distribution of the previous layer, and the marginal distribution for layer $\NumDGPlayer$ can be expressed using~\eqref{eq:svgp_marginal_qf}~and~\eqref{eq:svgp_marginal_qf_dist} as
\begin{align}\label{eq:DGP_marginal}
      \variationalprob(\cusvector{\processfunction}_{\numobservation}^{(\NumDGPlayer)}) = \int \prod^{\NumDGPlayer}_{\numDGPlayer=1} \variationalprob(\cusvector{\processfunction}_{\numobservation}^{(\numDGPlayer)})\text{d}\cusvector{\processfunction}_{\numobservation}^{(\numDGPlayer-1)},
\end{align}
where $\variationalprob(\cusvector{\processfunction}_{\numobservation}^{(\numDGPlayer)}) = \variationalprob(\cusvector{\processfunction}_{\numobservation}^{(\numDGPlayer)}\vert\cusvector{\processfunction}_{\numobservation}^{(\numDGPlayer-1)}, \cusmatrix{\Inducinglocations}^{(\numDGPlayer-1)})$ and $\cusvector{\trueouts}_{\numobservation}^{(0)}=\cusvector{\params}_{\numobservation}$. For scalar-valued observation, the latent function $\processfunction^{(\NumDGPlayer)}$ is a scalar, and the variational posterior can be approximated by minimizing the negative ELBO for DGPs using~\eqref{eq:DGP_marginal} as
\begin{align}\label{eq:DGP_ELBO}
      \mathcal{L}_{\substack{\text{DGP} \\ \text{ELBO}}} = -&\sum^{\Numobservation}_{\numobservation=1}\mathbb{E}_{\variationalprob\big(\processfunction_{\numobservation}^{(\NumDGPlayer)}\big)}\big[\log\probability(\obsouts_{\numobservation}\vert\processfunction_{\numobservation}^{(\NumDGPlayer)})\big] \nonumber\\+ \klconst&\sum^{\NumDGPlayer}_{\numDGPlayer=1}\text{KL}\big(\variationalprob(\cusmatrix{\Inducingpoints}^{(\numDGPlayer)})\vert\vert\probability(\cusmatrix{\Inducingpoints}^{(\numDGPlayer)})\big).
\end{align}
Since $\variationalprob(\cusvector{\processfunction}_{\numobservation}^{(\NumDGPlayer)})$ represents a distribution rather than a scalar value, an exact solution is intractable. \cite{salimbeni2017doubly} proposed the use of $\MCsamples$ Monte Carlo (MC) samples as, $\cusvector{\trueouts}^{(\numDGPlayer)}_{\numobservation, \mcsamples}\sim\variationalprob(\cusvector{\processfunction}_{\numobservation}^{(\numDGPlayer)}\vert\cusvector{\trueouts}_{\numobservation}^{(\numDGPlayer-1)}, \cusmatrix{\Inducinglocations}^{(\numDGPlayer-1)})$, to calculate the expected log probability.

\subsection{Gaussian Process Network and their limitations}\label{sec:gp_related_work}
Gaussian Process Networks (GPN) coined by~\cite{friedman2000gpn} and extended by \cite{giudice2024bayesian} addresses the learning the Bayesian network structure and not the inference, which is different from our work. Gaussian Process Regression Networks (GPRN) by~\cite{friedman2000gpn, wilson2011gaussian} provide a different perspective by combining modeling the final output as a linear combination of Gaussian processes (like neural network structure).
GPRN does not incorporate the intermediate observations and caters to a problem statement different from ours.

GPN introduced by~\cite{astudillo2021bayesian, aglietti2020causal} as the surrogate model would translate the toy process network shown in Figure~\ref{fig:multi_process} into the DAG shown in Figure~\ref{fig:toy_process_gpn}. The grey-shaded nodes represent the observed outputs, and the unshaded nodes represent the unobserved latent outputs. The so far introduced GPNs model each subprocess $\process^{(\numprocess)}$ as a Gaussian process with mean $\gpmean^{(\numprocess)}(\cdot)$ and variance $\gpkernel^{(\numprocess)}(\cdot, \cdot')$, where $\processfunction^{(\numprocess)}(\cdot)\sim\mathcal{GP}^{(\numprocess)}(\gpmean^{(\numprocess)}(\cdot),\gpkernel^{(\numprocess)}(\cdot, \cdot'))$ and the observation(s) of the subprocess $\process^{(\numprocess)}$ can be expressed using the respective Gaussian likelihood as $\probability(\obsouts^{\numprocess}\vert\processfunction^{(\numprocess)})$. The model assumes that each node GP is independent of the other GP given the observed input-output pairs and uses closed-form marginal log-likelihood (MLL)~\citep{rasmussen2003gaussian} for inference.
MC samples are used to estimate the final output during prediction. The implemented setup poses four major limitations:
\begin{enumerate}
      \item In most real-world cases, one can only observe the state space partially using an indirect observation lens. The latent outputs of the subprocesses are often hidden. Because of this, it can be considered that it is the latent/true outputs $\cusvector{\trueouts}^{(\numprocess)}$ that influence the process and not the indirect observations $\cusvector{\obsouts}^{(\numprocess)}$. Figure~\ref{fig:toy_process_gpn} represents that the indirect observations become the input to the respective child node. It holds only for direct noise-free observations.
      \item Using GP, the output of a parent node GP is a distribution, not a point value. Due to this reason, closed-form MLL cannot be used. Closed-form MLL can be used for deterministic inputs. Also, the prediction contradicts the training as MC samples are used to calculate from predictive distribution.
      \item The use of exact MLL also limits the usage to only Gaussian observation likelihood. Although this has computational benefits, it cannot be applied to cases where the intermediate observations are observed using a non-Gaussian lens.
      \item Using the inference method of existing GPN, one can only condition a particular node when observed. However, since the model is a network, one should be able to condition the connected subprocesses based on a particular subprocess observation.
\end{enumerate}

The models, demonstrated in~\cite{sussex2022model, kusakawa2022bayesian}, show promising results by overcoming the first limitation. However, they still rely on closed-form MLL for independent node inference, leaving the other limitations unaddressed. The models of~\cite{aglietti2020causal, sussex2022model, astudillo2021bayesian} were primarily proposed as a surrogate model for Bayesian optimization with intermediate observations, hinting at the potential for further development and improvement.

Another recently proposed variant of GPN is the Gaussian Process Autoregressive Regression model~\citep{requeima2019gaussian}, where the focus is on autoregressive modeling of each observed output. The prediction(s) of the previous output(s) are used as input(s) for the GP, which is further down the autoregressive flow. The outputs are ordered greedily, using an exhaustive search, but scalability is not well discussed. Although GPAR mentions the use of inducing points in D-GPAR-NL, it does not provide a training method for joint distribution loss and either assume fixed inducing locations or individual node GP training. Additionally, GPAR does not cater to the second and third limitations or provide an optimization method for inducing point formulation.

\section{Partially Observable Gaussian Process Network}\label{sec:pogpn}
We present our main contribution, the Partially Observable Gaussian Process Network (POGPN). Instead of node observations sharing a common distribution space, we propose that the latent functions reside in the same space and influence the child subprocess nodes. POGPN represents the process network in section~\ref{sec:multi_process} using a DAG where the nodes, $\numprocess\in\numprocessset$, are topologically ordered such that $\numprocess'<\numprocess,\forall \numprocess' \in \nodeparent{\numprocess}$. Each node is modeled as a $\mathcal{GP}^{(\numprocess)}$ similar to the GPN setup in section~\ref{sec:gp_related_work}, generalized as a vector-valued function $\cusvector{\trueouts^{(\numprocess)}}_{\numobservation}\sim\mathcal{GP}^{(\numprocess)}(\cdot, \cdot')$ and can or cannot be observed using an arbitrary likelihood $\probability(\cusvector{\obsouts}^{(\numprocess)}_{\numobservation}\vert\cusvector{\processfunction}^{(\numprocess)}_{\numobservation})$. The generalized notation allows for nodes to have different dimensionality. This formulation allows us to consider DGP a special POGPN case where only the last node observations are available. POGPN, with its assumption of arbitrary observation likelihood and common latent function space, provides a way to model continuous and categorical observations with the same model.

Unlike the existing GPNs, a subprocess $\process^{(\numprocess)}$ takes the parent node latent GP functions $\cusvector{\processfunction}^{\nodeparent{\numprocess}}$ as the input rather than an instance of the noisy indirect observations $\cusvector{\tilde{\obsouts}}^{\nodeparent{\numprocess}}$. Since we can express the node's latent function using a distribution, we can take the expectation over the parent node distribution. The expectation over possible parent node output(s) provides robustness against parent subprocess stochasticity and can separate the observation lens from the actual process. This setup also allows for arbitrary observation likelihoods as the observation is separated from the network. Using POGPN, the process network $\Process$ in Figure~\ref{fig:multi_process} would be represented as Figure~\ref{fig:toy_process_pogpn}.

\textbf{Evidence Lower BOund (ELBO).} Similar to DGP in section~\ref{sec:DGP}, we introduce inducing points, $\cusmatrix{\Inducinglocations}^{\nodeparent{\numprocess}}, \cusmatrix{\Inducinglocations}^{\paramsset^{(\numprocess)}}$, in the space of the parent nodes and node inputs respectively, such that $\cusmatrix{\Inducinglocations}^{(\numprocess)}=(\cusmatrix{\Inducinglocations}^{\nodeparent{\numprocess}}, \cusmatrix{\Inducinglocations}^{\paramsset^{(\numprocess)}})$. We wish to approximate the posterior $\probability\big(\{\cusvector{\processfunction}_{\numobservation}^{(\numprocess)}\}_{\numprocess\in\numprocessset}\vert\{\cusvector{\obsouts}_{\numobservation}^{(\numprocess)}; \cusvector{\params}_{\numobservation}^{(\numprocess)}\}_{\numprocess\in\numprocessset}\big)$ with the variational posterior $\variationalprob\big(\{\cusvector{\processfunction}_{\numobservation}^{(\numprocess)}\}_{\numprocess\in\numprocessset}\big)$ which can be expressed using~\eqref{eq:DGP_marginal} as
\begin{equation}\label{eq:POGPN_variational_marginal}
      \variationalprob \big(\{\cusvector{\processfunction}_{\numobservation}^{(\numprocess)}\}_{\numprocess\in\numprocessset}\big) = \prod_{\numprocess\in\numprocessset}\variationalprob\big(\cusvector{\processfunction}_{\numobservation}^{(\numprocess)};\{\cusvector{\processfunction}_{\numobservation}^{\nodeparent{\numprocess}},\cusvector{\params}_{\numobservation}^{(\numprocess)}\}, \cusmatrix{\Inducinglocations}^{(\numprocess)}\big).
\end{equation}
The Kullback Leibler (KL)~\citep{shlens2014notes} divergence between the variational posterior and true posterior can be expressed as
\begin{equation}
      \begin{gathered}\label{eq:POGPN_KL_variational}
            - \text{KL}\big(\variationalprob(\{\cusvector{\processfunction}_{\numobservation}^{(\numprocess)}\}_{\numprocess\in\numprocessset})\Vert\probability(\{\cusvector{\processfunction}_{\numobservation}^{(\numprocess)}\}_{\numprocess\in\numprocessset}\vert\{\cusvector{\obsouts}_{\numobservation}^{(\numprocess)},\cusvector{\params}_{\numobservation}^{(\numprocess)}\}_{\numprocess\in\numprocessset})\big) \\
            =\underbrace{\mathbb{E}_{\variationalprob(\{\cusvector{\processfunction}_{\numobservation}^{(\numprocess)}\}_{\numprocess\in\numprocessset})}\big[\log\probability(\{\cusvector{\obsouts}_{\numobservation}^{(\numprocess)}\}_{\numprocess\in\numprocessset}\vert\{\cusvector{\processfunction}_{\numobservation}^{(\numprocess)},\cusvector{\params}_{\numobservation}^{(\numprocess)}\}_{\numprocess\in\numprocessset})\big]}_{\text{Log Likelihood Loss (LL loss)}} \\ \underbrace{-\text{KL}\big(\variationalprob(\{\cusvector{\processfunction}_{\numobservation}^{(\numprocess)}\}_{\numprocess\in\numprocessset})\Vert\probability(\{\cusvector{\processfunction}_{\numobservation}^{(\numprocess)}\}_{\numprocess\in\numprocessset})\big)}_{\text{KL loss}} + \text{ Evidence}
      \end{gathered}
\end{equation}
The ELBO for POGPN can then be defined as the combination of the "LL loss" and the "KL loss" term of~\eqref{eq:POGPN_KL_variational}. We now show how the terms of the ELBO can be simplified so that inference can be performed.

The conditional distribution $\probability(\{\cusvector{\obsouts}_{\numobservation}^{(\numprocess)}\}_{\numprocess\in\numprocessset}\vert\{\cusvector{\processfunction}_{\numobservation}^{(\numprocess)}\}_{\numprocess\in\numprocessset})$ can be simplified using the DAG structure as
\begin{align}\label{eq:DAG_cond_independent}
       & \probability(\{\cusvector{\obsouts}_{\numobservation}^{(\numprocess)}\}_{\numprocess\in\numprocessset}\vert\{\cusvector{\processfunction}_{\numobservation}^{(\numprocess)}\}_{\numprocess\in\numprocessset})\nonumber                                                                                                                                                                        \\
       & = \probability(\cusvector{\obsouts}_{\numobservation}^{(\Numprocess)}\vert\{\cusvector{\processfunction}_{\numobservation}^{(\numprocess)}\}_{\numprocess\in\numprocessset})\probability(\cusvector{\obsouts}_{\numobservation}^{(\Numprocess-1)},\{\cusvector{\processfunction}_{\numobservation}^{(\numprocess)}\}_{\numprocess\in\numprocessset})\nonumber                                 \\
       & = \prod_{\numprocess\in\numprocessset}\probability(\cusvector{\obsouts}_{\numobservation}^{(\numprocess)}\vert\{\cusvector{\processfunction}_{\numobservation}^{(\numprocess)}\}_{\numprocess\in\numprocessset})= \prod_{\numprocess\in\numprocessset}\probability(\cusvector{\obsouts}_{\numobservation}^{(\numprocess)}\vert\cusvector{\processfunction}_{\numobservation}^{(\numprocess)})
\end{align}
where $\Numprocess=\vert\numprocessset\vert$. Using~\eqref{eq:POGPN_variational_marginal} and~\eqref{eq:DAG_cond_independent}, the "LL loss" in Equation\eqref{eq:POGPN_KL_variational}, can be expressed as
\begin{align}\label{eq:pogpn_elbo_ll}
        & \mathbb{E}_{\variationalprob(\{\cusvector{\processfunction}_{\numobservation}^{(\numprocess)}\}_{\numprocess\in\numprocessset})}\big[\log\probability(\{\cusvector{\obsouts}_{\numobservation}^{(\numprocess)}\}_{\numprocess\in\numprocessset}\vert\{\cusvector{\processfunction}_{\numobservation}^{(\numprocess)},\cusvector{\params}_{\numobservation}^{(\numprocess)}\}_{\numprocess\in\numprocessset})\big]\nonumber \\
      = & \mathbb{E}_{\variationalprob(\{\cusvector{\processfunction}_{\numobservation}^{(\numprocess)}\}_{\numprocess\in\numprocessset})}\big[\sum_{\numprocess\in\numprocessset}\log\probability(\cusvector{\obsouts}_{\numobservation}^{(\numprocess)}\vert\cusvector{\processfunction}_{\numobservation}^{(\numprocess)})\big]\nonumber                                                                                          \\
      = & \sum_{\numprocess\in\numprocessset} \mathbb{E}_{\variationalprob(\cusvector{\processfunction}_{\numobservation}^{(\numprocess)})}\big[\log\probability(\cusvector{\obsouts}_{\numobservation}^{(\numprocess)}\vert\cusvector{\processfunction}_{\numobservation}^{(\numprocess)})\big],
\end{align}
where the marginal $\variationalprob(\cusvector{\processfunction}_{\numobservation}^{(\numprocess)})$ can be calculated using~\eqref{eq:svgp_marginal_qf} and~\eqref{eq:svgp_marginal_qf_dist} as
\begin{align}\label{eq:POGPN_marginal}
      \variationalprob(\cusvector{\processfunction}_{\numobservation}^{(\numprocess)}) = \int\prod^{\numprocess}_{j\in\numprocessset}\variationalprob\big(\cusvector{\processfunction}_{\numobservation}^{(j)}\vert\cusvector{\processfunction}_{\numobservation}^{\nodeparent{j}}, \cusvector{\params}_{\numobservation}^{(j)}, \cusmatrix{\Inducinglocations}^{(j)}\big)\text{d}\cusvector{\processfunction}_{\numobservation}^{\nodeparent{j}}.
\end{align}
For $\Numobservation^{(\numprocess)}$ observations for each node $\numprocess$, the inference can be made by minimizing the negative EBLO for POGPN as
\begin{align}\label{eq:pogpn_elbo}
      \mathcal{L}_{\substack{\text{POGPN}                                                                                                                                                                                          \\\text{ELBO}}}^{(\numprocessset)} = &\sum_{\numprocess\in\numprocessset} \mathcal{L}_{\substack{\text{Node} \\ \text{ELBO}}} ^{(\numprocess)} \nonumber \\=&\underbrace{-\sum_{\numprocess\in\numprocessset}\frac{1}{\pogpnnormconst^{(\numprocess)}}\sum^{\Numobservation^{(\numprocess)}}_{\numobservation}\mathbb{E}_{\variationalprob(\cusvector{\processfunction}^{(\numprocess)}_{\numobservation})}\big[\log\probability(\cusvector{\obsouts}^{(\numprocess)}_{\numobservation}\vert\cusvector{\processfunction}^{(\numprocess)}_{\numobservation})\big]}_{\text{LL loss}}\nonumber \\
      + & \underbrace{\klconst\sum_{\numprocess\in\numprocessset}\text{KL}\big(\variationalprob(\cusvector{\inducingpoints}^{(\numprocess)})\Vert\probability(\cusvector{\inducingpoints}^{(\numprocess)})\big)}_{\text{KL loss}},
\end{align}
a normalization constant $\pogpnnormconst^{(\numprocess)}$ is introduced to keep the likelihood loss from different nodes comparable when the dimensionality of the node is not the same. We propose to keep $\pogpnnormconst^{(\numprocess)}=\dims{\obsouts^{(\numprocess)}}$ to keep the "LL loss" term of different nodes comparable and give equal importance to each node, where $\dims{\obsouts^{(\numprocess)}}$ is the dimension of the observed output $\obsouts^{(\numprocess)}$. However, $\pogpnnormconst^{(\numprocess)}$ incorporate importance-based training where the emphasis lies on a particular node. If a node $\numprocess'$ has no observation likelihood, then $\numprocess'$ will contribute to only the "KL loss" and not to the "LL loss".

\textbf{Predictive Log Likelihood (PLL) loss.} Using the inspiration from the PLL loss in~\ref{svgp:pll}~\citep{jankowiak2020parametric}, the "LL loss" for PLL can be defined for POGPN as
\begin{equation}\label{eq:pogpn_pll}
      \text{LL}_{\text{PLL}} = \underbrace{-\sum_{\numprocess\in\numprocessset}\frac{1}{\pogpnnormconst^{(\numprocess)}}\sum^{\Numobservation^{(\numprocess)}}_{\numobservation=1}\log\mathbb{E}_{\variationalprob(\cusvector{\processfunction}^{(\numprocess)}_{\numobservation})}\big[\probability(\cusvector{\obsouts}^{(\numprocess)}_{\numobservation}\vert\cusvector{\processfunction}^{(\numprocess)}_{\numobservation})\big]}_{\text{LL loss}},
\end{equation}
while the "KL loss" is the same as~\eqref{eq:pogpn_elbo}. Like DGP inference, we use MC samples to calculate the "LL loss" for POGPN ELBO in~\eqref{eq:pogpn_elbo}. It is common to calculate log probabilities to avoid loss of precision, and a direct summation of log probabilities of MC samples would lead to expected log marginal likelihood rather than log expected marginal likelihood in~\eqref{eq:pogpn_pll}. Using Jensen's inequality, one can prove that the former is an unbiased estimator of the latter. \cite{jankowiak2020deep} provides the sigma point method as a solution, but it is not easy to scale. We propose a more straightforward approach, where we use \verb|logsumexp| to calculate the log of expected likelihood marginal (LL loss) of PLL using MC samples as
\begin{align}\label{eq:mc_pogpn_pll_ll}
      \text{LL}_\text{PLL} & = -\sum_{\numprocess\in\numprocessset} \frac{1}{\pogpnnormconst^{(\numprocess)}} \sum^{\Numobservation^{(\numprocess)}}_{\numobservation=1} \log \Big(\frac{1}{\MCsamples}\sum^{\MCsamples}_{\mcsamples=1} \probability \big(\cusvector{\obsouts}^{(\numprocess)}_{\numobservation} \big| \cusvector{\trueouts}^{(\numprocess)}_{\numobservation, \mcsamples} \big)\Big)
\end{align} where $\cusvector{\trueouts}^{(\numprocess)}_{\numobservation, \mcsamples}\sim\variationalprob(\cusvector{\processfunction}_{\numobservation}^{(\numprocess)})$. This formulation has a tighter lower bound to the log expected marginal likelihood in comparison to the unbiased estimator. For generalization, we call ELBO and PLL the "loss" for POGPN. MC samples, used while training, can be considered analogous to training the child process on many hypothesized parent true/latent outputs, and the variational inference allows for robustness against the stochasticity of parent subprocesses.

We now present two methods, namely ancestor-wise and node-wise, for training POGPN . These methods can use either of the factorized losses, ELBO~\eqref{eq:pogpn_elbo} or PLL~\eqref{eq:pogpn_pll}. For $\Numprocess$ nodes, $\Numobservation$ observations, and $\Numinducing$ points for each node, the computational complexity is $\mathcal{O}(\Numprocess(\Numobservation\Numinducing^2+\Numinducing^3))$.

\textbf{Ancestor-wise Training.}
Algorithm~\ref{alg:pogpn_ancestorwise}  called POGPN-AL can be implemented using~\eqref{eq:pogpn_elbo} and~\eqref{eq:pogpn_pll}, where $\text{Anc}(\numprocessset_{\text{obs}})$ represents the set of all ancestors of each node $\numprocess\in\numprocessset_{\text{obs}}$; $\cusvector[bm]{\gphyperparam}^{(\numprocess)}$ represent the GP hyperparameters (mean, kernel and variational) and $\cusvector[bm]{\likelihoodparam}^{(\numprocess)}$ and likelihood hyperparameters of node $\numprocess$.
\begin{algorithm}
      \SetAlgoLined
      \caption{POGPN Ancestor-wise Loss (POGPN-AL) training. Given $\{\cusmatrix{\Obsouts}^{(\numprocess)}\}_{\numprocess\in\numprocessset_{\text{obs}}}$ observations for $\numprocessset_{\text{obs}}$,  GP hyperparameters of observed nodes $\cusvector[bm]\gphyperparam^{(\numprocessset_{\text{obs}})}$ and their ancestors $\cusvector[bm]\gphyperparam^{(\nodeancestor{\numprocessset_{\text{obs}}})}$ along with hyperparameters of observed likelihoods $\cusvector[bm]{\likelihoodparam}^{(\numprocessset_{\text{obs}})}$ are trained. One can use either ELBO or PLL loss from~\ref{eq:pogpn_elbo} or~\ref{eq:pogpn_pll} as $\mathcal{L}_{\text{POGPN}}$. $\cusvector[bm]{\gamma}^{(\numprocessset_{\text{obs}})}=(\cusvector[bm]{\gphyperparam}^{(\numprocessset_{\text{obs}})}, \cusvector[bm]{\gphyperparam}^{(\nodeancestor{\numprocessset_{\text{obs}}})},  \cusvector[bm]{\likelihoodparam}^{(\numprocessset_{\text{obs}})})$.}\label{alg:pogpn_ancestorwise}

      \SetKwInput{Input}{Input} 

      \Input{Training data: $\{\mathcal{D}^{(\numprocess)}\}_{\numprocess\in\numprocessset_{\text{obs}}} = \{\cusmatrix{\Obsouts}^{(\numprocess)}, \cusmatrix{\Params}^{(\numprocess)}\}_{\numprocess\in\numprocessset_{\text{obs}}}$}
      \SetInd{2em}{1em}
      \Indp 
      Loss: $\mathcal{L}_{\text{POGPN}}$\\
      Hyperparameters: $\cusvector[bm]{\gamma}^{(\numprocessset_{\text{obs}})}$\\
      Gradient optimizer: \texttt{optim}\\
      \Indm 
      \SetInd{1em}{1em}
      \While{not converged}{
      Compute $\variationalprob(\{\cusmatrix{\Processfunction}^{(\numprocess)}\}_{\numprocess \in \numprocessset_{\text{obs}}})$ using MC samples\;
      Compute $\mathcal{L}_{\text{POGPN}}(\numprocessset_{\text{obs}})$ using $\variationalprob(\{\cusmatrix{\Processfunction}^{(\numprocess)}\}_{\numprocess \in \numprocessset_{\text{obs}}})$ and $\{\mathcal{D}^{(\numprocess)}\}_{\numprocess\in\numprocessset_{\text{obs}}}$\;
      Gradient step: $\cusvector[bm]{\gamma}^{(\numprocessset_{\text{obs}})} \leftarrow \texttt{optim}(\mathcal{L}_{\text{POGPN}}^{(\numprocessset_{\text{obs}})})$\;
      }
      \KwOut{Optimized hyperparameters: $\cusvector[bm]{\gamma}^{(\numprocessset_{\text{obs}})}$}
\end{algorithm}
We call Algorithm~\ref{alg:pogpn_ancestorwise} ancestor-wise training as it updates the parameters of all ancestor GPs of the observed nodes. This method is similar to the traditional method of training DGP, just that we consider multiple observation nodes in POPGN. It is beneficial when either all network nodes are observed or the nodes further in the graph are observed, and one wishes to condition the ancestor node(s) based on the observations of the child/successor node(s).

\begin{figure}[h]
      \centering
      \tikzset{every picture/.style={line width=0.7pt}} 
\begin{tikzpicture}[node distance=30mm and 10mm,
        process/.style={circle, draw, minimum size=6mm, align=center, inner sep=0.5pt},
        every node/.style={font=\footnotesize},
        arr/.style={-Stealth},
        obs/.style={circle, draw, fill=gray!30, minimum size=6mm, align=center, inner sep=0.5pt}
    ]
    \node[process] (f1) {$\cusvector{\processfunction}^{(1)}$};
    \node[obs] (x1) [above = of f1,yshift=-2.5cm, xshift=-1.0cm ] {$\cusvector{\params}^{(1)}$};
    \node[process] (y1) [above=of f1, yshift=-2.5cm] {$\cusvector{\obsouts}^{(1)}$};
    \node[process] (f2) [right= of f1, xshift=1.2cm]{$\cusvector{\processfunction}^{(2)}$};
    \node[process] (y2) [above=of f2, yshift=-2.5cm] {$\cusvector{\obsouts}^{(2)}$};
    \node[process] (f3) [right= of f2, xshift=1.0cm]{$\cusvector{\processfunction}^{(3)}$};
    \node[process] (y3) [above=of f3, yshift=-2.5cm] {$\cusvector{\obsouts}^{(3)}$};
    \node[process] (f4) [below= of f1, yshift=2.5cm, xshift=1.0cm]{$\cusvector{\processfunction}^{(4)}$};
    \node[obs] (y4) [left=of f4, xshift=-0.3cm] {$\cusvector{\obsouts}^{(4)}$};

    \node[process] (f5) [right= of f4]{$\cusvector{\processfunction}^{(5)}$};
    \node[obs] (y5) [right=of f5, xshift=0.0cm] {$\cusvector{\obsouts}^{(5)}$};

    \draw[arr] (f1) -- (y1);
    \draw[arr] (x1) -- (f1);
    \draw[arr] (f2) -- (y2);
    \draw[arr] (f3) -- (y3);
    \draw[arr] (f4) -- (y4);
    \draw[arr] (f5) -- (y5);

    \draw[arr] (f1) -- (f2);
    \draw[arr] (f2) -- (f3);
    \draw[arr] (f2) -- (f5);
    \draw[arr] (f1) -- (f4);
    \draw[arr] (f4) -- (f5);

    \plate [dashed, inner sep=0.15cm, draw=blue] {AL} {(f1) (f2) (f4) (f5) (y4) (y5)} {};
    \plate [dashed, inner sep=0.05cm, draw=red] {NL} {(f4) (y4)} {};
    \plate [dashed, inner sep=0.05cm, draw=red] {NL} {(f5) (y5)} {};

\end{tikzpicture}
      \caption{Training methods POGPN for a given structure. If $\numprocessset_{\text{obs}}=\{\cusvector{\obsouts}^{(4)}, \cusvector{\obsouts}^{(5)}\}$, POGPN-AL includes hyperparameters for node $\cusvector[bm]{\gamma}^{(\numprocessset_{\text{obs}})}=(\cusvector[bm]{\gphyperparam}^{(\numprocessset_{\text{obs}})}, \cusvector[bm]{\gphyperparam}^{(\nodeancestor{\numprocessset_{\text{obs}}})},  \cusvector[bm]{\likelihoodparam}^{(\numprocessset_{\text{obs}})})$ bounded by the blue dashed box. POGPN-NL trains hyperparameters, $\cusvector[bm]{\gamma}^{(\numprocess)} = (\cusvector[bm]{\gphyperparam}^{(\numprocess)}, \cusvector[bm]{\likelihoodparam}^{(\numprocess)}), \forall\numprocess\in\numprocessset_{\text{obs}}$, node-wise as bounded by red dashed boxes. Gray nodes represent observed output nodes (likelihood), and white nodes represent latent output nodes (GP).}
      \vspace{-1em}
      \label{fig:training}
\end{figure}

\begin{algorithm}
      \SetAlgoLined
      \caption{POGPN Node-wise Loss (POGPN-NL) training. Given $\{\Numobservation^{(\numprocess)}\}_{\numprocess\in\numprocessset_{\text{obs}}}$ observations for $\numprocessset_{\text{obs}}$, the GP hyperparameters $\cusvector[bm]\gphyperparam^{(\numprocess)}$ and likelihood hyperparameters $\cusvector[bm]{\likelihoodparam}^{(\numprocess)}$ are trained for one node at a time for $\numprocess\in\numprocessset_{\text{obs}}$. One can use either ELBO or PLL loss from~\ref{eq:pogpn_elbo} or~\ref{eq:pogpn_pll} as $\mathcal{L}_{\text{POGPN}}$. $\cusvector[bm]{\gamma}^{(\numprocessset_{\text{obs}})}=(\cusvector[bm]{\gphyperparam}^{(\numprocessset_{\text{obs}})}, \cusvector[bm]{\likelihoodparam}^{(\numprocessset_{\text{obs}})})$.}\label{alg:pogpn_node-wise}

      \SetKwInput{Input}{Input} 

      \Input{Training data: $\{\mathcal{D}^{(\numprocess)}\}_{\numprocess\in\numprocessset_{\text{obs}}} = \{\cusmatrix{\Obsouts}^{(\numprocess)}, \cusmatrix{\Params}^{(\numprocess)}\}_{\numprocess\in\numprocessset_{\text{obs}}}$}
      \SetInd{2em}{1em}
      \Indp 
      Loss: $\mathcal{L}_{\text{POGPN}}$\\
      Gradient optimizer: \texttt{optim}\\
      \Indm 
      \SetInd{1em}{1em}

      \While{not converged}{
      \For{$\numprocess \in \text{\texttt{topological\_sort}}(\numprocessset_{\text{obs}})$}{
      Hyperparameters: $\cusvector[bm]{\gamma}^{(\numprocess)} = (\cusvector[bm]{\gphyperparam}^{(\numprocess)}, \cusvector[bm]{\likelihoodparam}^{(\numprocess)})$\\
      Compute $\variationalprob(\cusmatrix{\Processfunction}^{(\numprocess)})$ using MC samples\;
      Compute $\mathcal{L}_{\text{node}}^{(\numprocess)}$ using $\variationalprob(\cusmatrix{\Processfunction}^{(\numprocess)})$ and  $\mathcal{D}^{(\numprocess)}$\;
      Gradient step $\cusvector[bm]{\gamma}^{(\numprocess)} \leftarrow \texttt{optim}(\mathcal{L}_{\text{node}}^{(\numprocess)})$\;
      }
      }
      \KwOut{Optimized hyperparameters: $\cusvector[bm]{\gamma}^{(\numprocessset_{\text{obs}})}$}
\end{algorithm}

\textbf{Node-wise Training.} Algorithm~\ref{alg:pogpn_node-wise}  called POGPN-NL, follows a coordinate ascent method for updating individual node GP hyperparameters $\cusvector[bm]{\gphyperparam}^{(\numprocess)}$ and likelihood hyperparameters $\cusvector[bm]{\likelihoodparam}^{(\numprocess)}$ for $\numprocess\in\numprocessset_{\text{obs}}$. With experimentation, we found that calculating updated $\variationalprob(\cusmatrix{\Processfunction}^{(\numprocess)})$ by looping over the observed nodes helps node-wise training converge to a global minimum. This is not the case when $\variationalprob(\cusmatrix{\Processfunction}^{(\numprocess)})$ is calculated only once outside the loop over nodes. Algorithm~\ref{alg:pogpn_node-wise} explains the node-wise coordinate ascent method.

\section{Experiments}\label{sec:experiments}
In this section, we conduct a comprehensive comparison of the performance of POGPN with various models, including independent GPs (IGP), Semi-Parameteric Latent Factor Model (SLFM), GPRN~\cite{wilson2011gaussian} and GPAR~\cite{requeima2019gaussian}.
POGPN is implemented using the gpytorch package~\cite{gardner2018gpytorch}. Here, we use a squared exponential kernel and constant mean for all experiments, as in other models. Similarly, we take the number of inducing locations the same as used by the D-GPAR-NL model from~\cite{requeima2019gaussian} to ensure proper comparison. We use the ICM variational approximation to model the multi-task nodes as proposed by~\cite{van2020framework}. The detailed construction procedure of POGPNs has been included in the supplementary section.

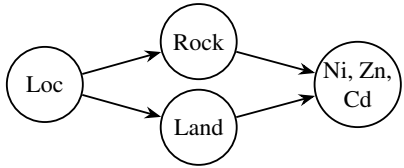
\begin{figure}[h]
      \centering
      \tikzset{every picture/.style={line width=0.7pt}} 
\begin{tikzpicture}[node distance=10mm and 10mm,
        process/.style={circle, draw, minimum size=10mm, align=center, inner sep=0.5pt},
        every node/.style={font=\footnotesize},
        arr/.style={-Stealth},
        obs/.style={circle, draw, fill=gray!30, minimum size=10mm, align=center, inner sep=0.5pt}
    ]
    \node[process] (loc) {Loc};
    \node[process] (rock) [right=of loc, yshift=0.57cm] {Rock};
    \node[process, draw=none] (mid_point) [right=of loc] {};
    \node[process] (land) [right=of loc, yshift=-0.57cm] {Land};
    \node[process] (cd) [right=of mid_point] {Ni, Zn,\\Cd};

    \draw[arr] (loc) -- (rock);
    \draw[arr] (loc) -- (land);
    \draw[arr] (land) -- (cd);
    \draw[arr] (rock) -- (cd);

\end{tikzpicture}
      \caption{Structure of POGPN with root node location "Loc", softmax likelihoods for "Rock" and "Land", and multi-task Gaussian likelihood for minerals.}
      \vspace{-1em}
      \label{fig:jura_dag}
\end{figure}

\textbf{Jura dataset}\footnote{The dataset can be downloaded from https://r-spatial.github.io/gstat/reference/jura.html.}. There are 259 locations from a mining area, for which the amount of zinc, nickel, and cadmium found is given. Along with these locations, there are another 100 locations with only zinc and nickel values available. The existing experiments use this information to predict the amount of cadmium for the remaining 100 locations. However, the original dataset also records two categorical observations: land use (4 classes) and the type of rock (5 classes) found at every location. Since all previous models cannot do classification and regression using one model, they do not use this information. However, we make a POGPN, as shown in Fig.~\ref{fig:jura_dag}, that can also use the categorical observations to predict the final output. We use softmax likelihood to model the multi-class observations and multi-task Gaussian likelihood (using LMC) to model mineral observations. For the latent function, we assume a two-dimensional multi-task GP as the latent function for categorical nodes "Rock" and "Land" and a three-dimensional GP for regression node "Zn, Ni, Cd." A detailed explanation of the POGPN structure has been shown in the supplementary section.

\begin{table}[h]
      \centering
      \begin{tabular}{@{}c@{}}
            \begin{tabular}{@{\hskip 0pt}lcccc@{\hskip 0pt}}
                  \toprule
                  Model & IGP\textsuperscript{\textdagger} & SLFM\textsuperscript{\textdagger} & GPRN\textsuperscript{\textdagger} & D-GPAR-NL\textsuperscript{\textdagger} \\
                  \midrule
                  MAE   & 0.5753                           & 0.4145                            & 0.4040                            & 0.3996                                 \\
                  \bottomrule
            \end{tabular}
            \\[2em] 

            \begin{tabular}{@{\hskip 0pt}lccc@{\hskip 0pt}}
                  \toprule
                  Model & POGPN-AL\textsuperscript{\textdaggerdbl} & POGPN-NL\textsuperscript{\textdaggerdbl} & POGPN-AL\textsuperscript{\textasteriskcentered} \\
                  \midrule
                  MAE   & 0.3991
                        & \textbf{0.3989}                          & 0.5035                                                                                     \\
                  \bottomrule
            \end{tabular}
      \end{tabular}
      \caption{Prediction results for Jura dataset. Mean absolute error (MAE) (lower is better). Models marked with \textsuperscript{\textdagger} indicate cited results from~\cite{requeima2019gaussian}. POGPN-AL\textsuperscript{\textdaggerdbl} and POGPN-NL\textsuperscript{\textdaggerdbl} are calculated using PLL. POGPN-AL\textsuperscript{\textasteriskcentered} is calculated using ELBO.}
      \vspace{-1em}
\end{table}

The number of inducing locations is 259, equal to the number of locations for fully observed data. The values are log standardized for evaluation, used for training, and then transformed back, and the mean absolute error is calculated. It can be seen that POGPN outperforms all other models. This shows POGPN as a new state-of-the-art multi-task that can even use multimodal intermediate information.

\begin{figure}[h]
      \centering
      \begin{subfigure}[t]{\columnwidth}
            \centering
            \tikzset{every picture/.style={line width=0.7pt}} 
\begin{tikzpicture}[node distance=10mm and 10mm,
        process/.style={circle, draw, minimum size=10mm, align=center, inner sep=0.5pt},
        every node/.style={font=\footnotesize},
        arr/.style={-Stealth},
        obs/.style={circle, draw, fill=gray!30, minimum size=6.5mm, align=center, inner sep=0.5pt}
    ]
    \node[process] (time) {Time};
    \node[process] (f3_f4_f5_f6) [right=of time] {F3, F4,\\F5, F6};
    \node[process] (f1_f2_fz) [right=of f3_f4_f5_f6] {F1, F2, \\FZ};

    \draw[arr] (time) -- (f3_f4_f5_f6);
    \draw[arr] (f3_f4_f5_f6) -- (f1_f2_fz);

\end{tikzpicture}
            \caption{POGPN for EEG dataset}
            \label{fig:eeg_dag}
      \end{subfigure}
      \hfill
      \begin{subfigure}[t]{\columnwidth}
            \centering
            \input{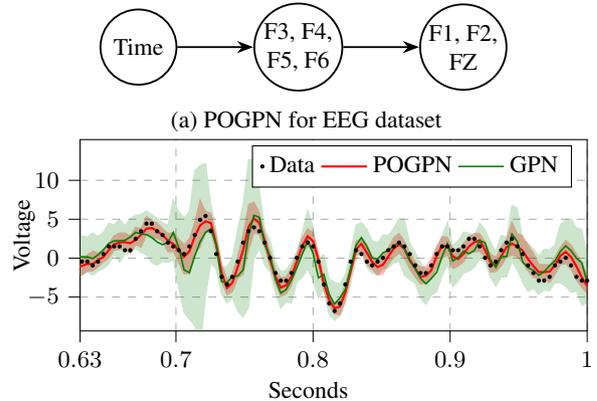}
            \caption{Prediction results for F2 sensor from EEG dataset.}
            \label{fig:eeg_pred}
      \end{subfigure}
      \caption{Structure of POGPN for EEG dataset in Figure~\ref{fig:eeg_dag}. Figure~\ref{fig:eeg_pred} shows prediction results for sensor F2 using POGPN-AL (PLL) and GPN with a similar concept as described by Figure~\ref{fig:toy_process_gpn}.} 
      \label{fig:eeg_combined} 
      \vspace{-1em}
\end{figure}

\textbf{EEG dataset}\footnote{The dataset can be downloaded from https://archive.ics.uci.edu/dataset/121/eeg+database.}. The dataset consists of electrode measurements from the scalp of different subjects. Each sensor records 256 voltage measurements. The data focuses on the measurements from sensors F1, F2, F3, F4, F5, F6, and FZ from the first trial of control subject 337. The task is to predict F1, F2, and FZ measurements for the last 100 timestep, given the full observation of F3, F4, F5, and F6, and the first 156 measurements of F1, F2, and FZ. The values are standardized before training. We make a POGPN as shown in Fig.~\ref{fig:eeg_dag}, where the intermediate node is a four-dimensional multi-task GP, and the final node is a three-dimensional multi-task GP with respective multi-task Gaussian likelihoods.

\begin{table}[h]
      \setlength{\tabcolsep}{4pt}
      \centering
      \begin{tabular}{@{}c@{}}
            \begin{tabular}{@{\hskip 0pt}lcccc@{\hskip 0pt}}
                  \toprule
                  Model & IGP\textsuperscript{\textdagger} & SLFM\textsuperscript{\textdagger} & GPAR-NL\textsuperscript{\textdagger} & POGPN-AL (PLL) \\
                  \midrule
                  SMSE  &
                  1.75  &
                  1.06  &
                  0.26  &
                  \textbf{0.24}
                  \\
                  MLL   &
                  2.60  & 4.00                             & 1.63                              & 1.04
                  \\
                  \bottomrule
            \end{tabular}
            \\[2em] 

            \begin{tabular}{@{\hskip 0pt}lcc@{\hskip 0pt}}
                  \toprule
                  Model & POGPN-NL (PLL) & POGPN-AL (ELBO) \\
                  \midrule
                  SMSE  & 0.28           & 0.31            \\
                  MLL   & \textbf{0.18}  & 1.40            \\
                  \bottomrule
            \end{tabular}
      \end{tabular}
      \caption{Prediction results for EEG dataset of different models. Standardized mean squared error (SMSE) and Mean Log Loss (MLL)~\cite{rasmussen2003gaussian} comparison (lower is better). Models marked with \textsuperscript{\textdagger} indicate cited results from~\cite{requeima2019gaussian}.}
      \vspace{-1em}
\end{table}
Inducing locations are in the "Time" domain and kept the same as the total time steps (256 points) as used by~\cite{requeima2019gaussian} and remain constant throughout the training process. For evaluation, the values are standardized before training and transformed back before prediction evaluation. POGPN consistently outperforms GPAR, showing significant improvements in SMSE and MLL. The results demonstrate the robustness of POGN against process stochasticity and the potential for even better confidence intervals, as shown in Figure~\ref{fig:eeg_pred}.

\textbf{Synthetic experiment.} We use the synthetic experiment from~\cite{requeima2019gaussian} but change it to have categorical observations to test the performance of POGPN for non-Gaussian noise along with categorical and continuous observations from subprocesses. Since the existing GPNs cannot incorporate categorical intermediate observations, we compare against DGP with the same layer structure but without intermediate subprocess likelihoods. For $x\in[0, 1]$, the system is described as
\begin{align*}
      f_{1}(x)
       & = -\frac{\sin\!\bigl(10\pi\,(x + 1)\bigr)}{2x + 1} - x^{4}           \\
      f_{2}(x)
       & = \cos^{2}\!\bigl(f_{1}(x)\bigr) + \sin(3x),                         \\
      f_{3}(x)
       & =\begin{cases}
                f_{2}(x)\,f_{1}(x)^{2} + 3\,x - 2.5, & \text{if } f_{2}(x) < 1.5,   \\
                f_{2}(x)\,f_{1}(x)^{2} + 3\,x + 2.5, & \text{if } f_{2}(x) \ge 1.5,
          \end{cases}
\end{align*}
and the intermediate observations are defined as
\begin{align*}
      y_{1} & = f_{1}(x) + \delta_{\text{obs}}, \\
      y_{2} & =
      \begin{cases}
            0, & \text{if } f_{2}(x) < 1.5,   \\
            1, & \text{if } f_{2}(x) \ge 1.5,
      \end{cases}
\end{align*}
where final output $y_3=f_3 + \delta_{\text{obs}}$ and $\delta_{\text{obs}} \sim \mathcal{N}(0, 0.1)$.
\begin{figure}[ht]
      \centering
      \begin{subfigure}[t]{\columnwidth}
            \centering


\begin{tikzpicture}[
        node distance=8mm and 8mm,       
        process/.style={circle, draw, minimum size=8mm, align=center, inner sep=0.5pt},
        every node/.style={font=\scriptsize},     
        arr/.style={-Stealth}
    ]

    \node[process] (x) {$x$};
    \node[process, right=of x] (f1) {$f^{(1)}$};
    \node[process, right=of f1] (f2) {$f^{(2)}$};
    \node[process, right=of f2] (f3) {$f^{(3)}$};

    \draw[arr] (x) -- (f1);
    \draw[arr] (f1) -- (f2);
    \draw[arr] (f2) -- (f3);

    \draw[arr, bend left=23]  (x) to (f2);
    \draw[arr, bend left=25]  (x) to (f3);
    \draw[arr, bend right=23] (f1) to (f3);

\end{tikzpicture}
            \caption{Structure of POGPN for synthetic experiment with Gaussian likelihood for $f^{(1)}$ and $f^{(3)}$, and Bernoulli likelihood for $f^{(2)}$.}
            \label{fig:synthetic_dag}
      \end{subfigure}
      \hfill
      \begin{subfigure}[t]{\columnwidth}
            \centering
            \input{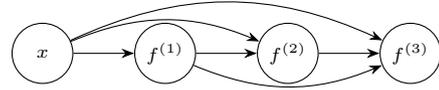}
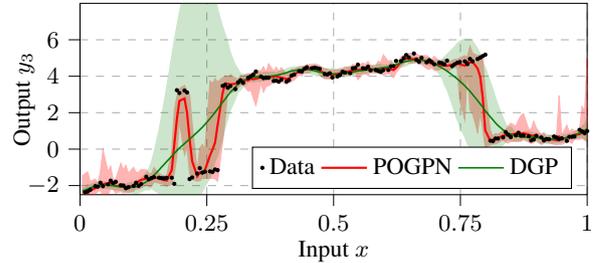
            \caption{Prediction results for $y_3$ for synthetic experiment.}
            \label{fig:synthetic_pred}
      \end{subfigure}
      \caption{Structure of POGPN for synthetic experiment in Figure~\ref{fig:synthetic_dag}. Figure~\ref{fig:synthetic_pred} shows prediction results for output $y_3$ test data using POGPN-AL and PLL loss with $\klconst=0.5$. Deep GP has the same hierarchical structure as GPs.} 
      \label{fig:synthetic_combined} 
      \vspace{-1em}
\end{figure}

We train on 40 equally spaced inputs in the range of $x$, and the number of inducing points is the same as training data points. As shown in Figure~\ref{fig:synthetic_pred}, POGPN can incorporate categorical intermediate subprocess observations, learn complex non-Gaussian process structure, and is robust against noise. The DGP cannot learn the process structure and provides wide confidence intervals due to high uncertainty, which has been discussed in detail by~\cite{duvenaud2014avoiding}. This also states the benefit of POGPN: intermediate node likelihoods can help the hierarchical GP model learn process structure better, thereby eliminating the pathology of deep networks and improving the predictive performance.

\section{Discussion}\label{sec:discussion}
We propose a Partially Observable Gaussian Process Network (POGPN) to model process networks where subprocesses can be stochastic, and the intermediate outputs can be observed using an observation lens, modeled as the observation likelihood. We develop a trainable loss, namely ELBO (Evidence Lower BOund) and Predictive Log Likelihood (PLL), for POGPN that makes inferences on the joint distribution of latent space and not just independent sub-processes. The inference can be made using MC samples, which can be considered analogous to training the child process on many hypothesized parent true/latent outputs. POGPN can incorporate continuous observations and non-Gaussian observations like categorical data. In our experience, we have not found any Gaussian process framework encompassing regression and classification within a single model. This setup makes POGPN very versatile and close to the real-world process networks where subprocesses can have arbitrary likelihood. We propose an ancestor-wise and a node-wise training method for POGPN. Experiments show the superior performance of POGPN-PLL over other existing Gaussian process networks. For further research, one can implement a message-passing algorithm to accommodate for parallel inference or use more complex observation likelihoods like a neural network.

\vspace{-1em}
\section*{Acknowledgement}
The research in this paper was supported by the Deutsche Forschungsgemeinschaft (DFG, German Research Foundation) - 459291153 (FOR5399).

\bibliographystyle{abbrvnat}
\bibliography{references}

\newpage

\onecolumn

\title{Partially Observable Gaussian Process Network and Doubly Stochastic Variational Inference\\(Supplementary Material)}
\maketitle
\appendix
\section{Experimental setup of Jura dataset}
\begin{figure}[h]
      \centering
      \tikzset{every picture/.style={line width=0.7pt}} 
\begin{tikzpicture}[node distance=10mm and 10mm,
        process/.style={circle, draw, minimum size=10mm, align=center, inner sep=0.5pt},
        every node/.style={font=\footnotesize},
        arr/.style={-Stealth},
        obs/.style={circle, draw, fill=gray!30, minimum size=10mm, align=center, inner sep=0.5pt}
    ]
    \node[process] (loc) {Loc};
    \node[process] (rock) [right=of loc, yshift=0.57cm] {$\cusvector{\processfunction}^{(\text{Rock})}$};
    \node[obs] (y_rock) [above=of rock, yshift=-0.5cm] {$\cusvector{\obsouts}^{(\text{Rock})}$};
    \node[process, draw=none] (mid_point) [right=of loc] {};
    \node[process] (land) [right=of loc, yshift=-0.57cm] {$\cusvector{\processfunction}^{(\text{Land})}$};
    \node[obs] (y_land) [below=of land, yshift=0.5cm] {$\cusvector{\obsouts}^{(\text{Land})}$};
    \node[process] (cd) [right=of mid_point] {$\cusvector{\processfunction}^{(\text{Ni, Zn, Cd})}$};
    \node[obs] (y_cd) [right=of cd] {$\cusvector{\obsouts}^{(\text{Ni, Zn, Cd})}$};

    \draw[arr] (loc) -- (rock);
    \draw[arr] (loc) -- (land);
    \draw[arr] (land) -- (cd);
    \draw[arr] (rock) -- (cd);
    \draw[arr] (rock) -- (y_rock);
    \draw[arr] (land) -- (y_land);
    \draw[arr] (cd) -- (y_cd);

\end{tikzpicture}
      \caption{Detailed structure of POGPN with root node location "Loc", softmax likelihoods for "Rock" and "Land" and multi-task Gaussian likelihood for minerals. Gray nodes represent observed nodes, and white nodes represent latent nodes.}
      \vspace{-1em}
      \label{fig:jura_dag_appendix}
\end{figure}
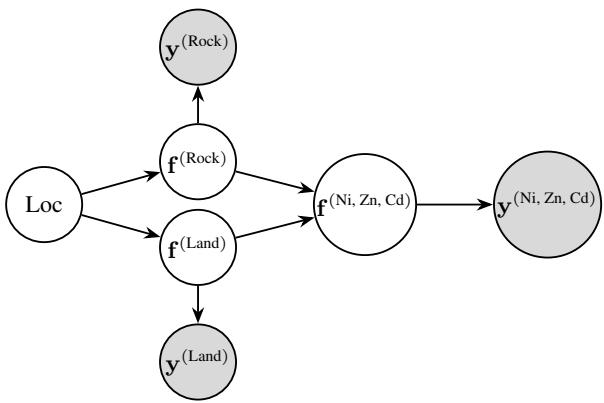

We make a POGPN, as shown in Fig.~\ref{fig:jura_dag_appendix}, that can also use the categorical observations to predict the final output. The values are log standardized before modeling and then transformed back for evaluation.

Latent functions $\cusvector{\processfunction}^{(\text{Rock})}$ and $\cusvector{\processfunction}^{(\text{Land})}$ are vector-valued GP of size two where the LMC kernel is used. $\cusvector{\processfunction}^{(\text{Ni, Zn, Cd})}$ is a vector-valued GP of size three where the LMC kernel is used. As mentioned in the main document, softmax likelihood is used for the "Rock" and "Land" nodes, and multi-task Gaussian likelihood is used for "Ni, Zn, Cd". The number of inducing locations is initialized as the fully observed 259 input locations. $\beta=2.5$ for the loss functions PLL and ELBO. For the fully observed dataset where "Rock" type, "Land" type, and "Zn, Ni, Cd" are observed for 156 locations, the number of epochs = 200. For the rest of the partially observed dataset of only "Rock" type, "Land" type, and "Zn, Ni," the number of epochs = 50. A squared exponential kernel and constant mean were used for all the kernels. For the test locations, the predicted multi-variate normal is conditioned on the known locations of Zn and Ni to get a normal distribution of Cd. We use ADAM~\citep{kingma2014adam} optimizer from PyTorch, with a learning rate of 0.01, to optimize the loss function.

\begin{table}[h]
      \centering
      \begin{tabular}{@{}c@{}}
            \begin{tabular}{@{\hskip 0pt}lcccc@{\hskip 0pt}}
                  \toprule
                  Model & IGP\textsuperscript{\textdagger} & SLFM\textsuperscript{\textdagger} & GPRN\textsuperscript{\textdagger} & D-GPAR-NL\textsuperscript{\textdagger} \\
                  \midrule
                  MAE   & 0.5753                           & 0.4145                            & 0.4040                            & 0.3996                                 \\
                  \bottomrule
            \end{tabular}
            \\[2em] 

            \begin{tabular}{@{\hskip 0pt}lccc@{\hskip 0pt}}
                  \toprule
                  Model & POGPN-AL\textsuperscript{\textdaggerdbl} & POGPN-NL\textsuperscript{\textdaggerdbl} & POGPN-AL\textsuperscript{\textasteriskcentered} \\
                  \midrule
                  MAE   & 0.3991$\pm$0.003
                        & \textbf{0.3989}$\pm$\textbf{0.002}       & 0.5035$\pm$0.0012                                                                          \\
                  \bottomrule
            \end{tabular}
      \end{tabular}
      \caption{Prediction results for Jura dataset. Mean absolute error (MAE) (lower is better). Models marked with \textsuperscript{\textdagger} indicate cited results from~\cite{requeima2019gaussian}. POGPN-AL\textsuperscript{\textdaggerdbl} and POGPN-NL\textsuperscript{\textdaggerdbl} are calculated using PLL. POGPN-AL\textsuperscript{\textasteriskcentered} is calculated using ELBO.}
      \vspace{-1em}
\end{table}

\section{Experimental setup of EEG dataset}

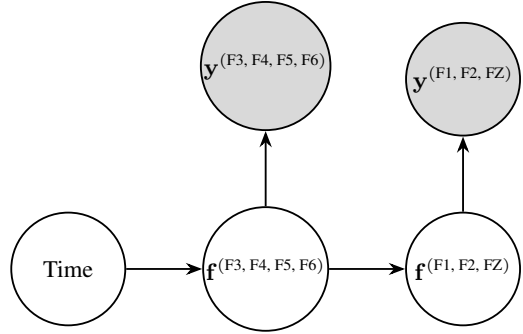
\begin{figure}[h]
      \centering
      \tikzset{every picture/.style={line width=0.7pt}} 
\begin{tikzpicture}[node distance=10mm and 10mm,
        process/.style={circle, draw, minimum size=15mm, align=center, inner sep=0.5pt},
        every node/.style={font=\footnotesize},
        arr/.style={-Stealth},
        obs/.style={circle, draw, fill=gray!30, minimum size=15mm, align=center, inner sep=0.5pt}
    ]
    \node[process] (time) {Time};
    \node[process] (f3_f4_f5_f6) [right=of time] {$\cusvector{\processfunction}^{(\text{F3, F4, F5, F6})}$};
    \node[obs] (y_f3_f4_f5_f6) [above=of f3_f4_f5_f6] {$\cusvector{\obsouts}^{(\text{F3, F4, F5, F6})}$};

    \node[process] (f1_f2_fz) [right=of f3_f4_f5_f6] {$\cusvector{\processfunction}^{(\text{F1, F2, FZ})}$};
    \node[obs] (y_f1_f2_fz) [above=of f1_f2_fz] {$\cusvector{\obsouts}^{(\text{F1, F2, FZ})}$};

    \draw[arr] (time) -- (f3_f4_f5_f6);
    \draw[arr] (f3_f4_f5_f6) -- (f1_f2_fz);
    \draw[arr] (f3_f4_f5_f6) -- (y_f3_f4_f5_f6);
    \draw[arr] (f1_f2_fz) -- (y_f1_f2_fz);

\end{tikzpicture}
      \caption{Detailed structure of POGPN with root node time, and multi-task Gaussian likelihoods for "F3, F4, F5 and F6" and "F1, F2 and FZ". Gray nodes represent observed nodes, and white nodes represent latent nodes.}
      \label{fig:eeg_dag_appendix}
\end{figure}

We make a POGPN as shown in Figure~\ref{fig:eeg_dag_appendix}, where the values are standardized before training and then back-standardized before prediction. The number of inducing locations for the "Time" node is initialized as the 256-time steps and is kept non-learnable. For the fully observed dataset where F1, F2, F3, F4, F5, F6 and FZ are observed for 156 timesteps, number of epochs = 300 and for the partially observed dataset of only F3, F4, F5 and F6 number of epochs = 150. For all kernels, squared exponential kernel and constant mean have been used, and the number of MC samples = 20. Latent functions $\cusvector{\processfunction}^{(\text{F3, F4, F5, F6})}$ and $\cusvector{\processfunction}^{(\text{F1, F2, FZ})}$ are four and three dimensional respectively. For each of the latent functions, the likelihood is multi-task Gaussian likelihood. We use ADAM~\citep{kingma2014adam} optimizer from PyTorch, with a learning rate of 0.02, to optimize the loss function. The detailed prediction results for the EEG dataset are explained in Table~\ref{tb:eeg_results_extended}.

\begin{table}[h]
      \setlength{\tabcolsep}{4pt}
      \centering
      \begin{tabular}{@{}c@{}}
            \begin{tabular}{@{\hskip 0pt}lcccc@{\hskip 0pt}}
                  \toprule
                  Model & IGP\textsuperscript{\textdagger} & SLFM\textsuperscript{\textdagger} & GPAR-NL\textsuperscript{\textdagger} & POGPN-AL (PLL) \\
                  \midrule
                  SMSE  &
                  1.75  &
                  1.06  &
                  0.26  &
                  \textbf{0.24}
                  $\pm$\textbf{0.016}
                  \\
                  MLL   &
                  2.60  & 4.00                             & 1.63                              & 1.04
                  $\pm$0.11
                  \\
                  \bottomrule
            \end{tabular}
            \\[2em] 

            \begin{tabular}{@{\hskip 0pt}lccc@{\hskip 0pt}}
                  \toprule
                  Model & POGPN-NL (PLL)                  & POGPN-AL (ELBO) & POGPN-NL (ELBO) \\
                  \midrule
                  SMSE  & 0.28$\pm$0.02                   & 0.31            & 0.34            \\
                  MLL   & \textbf{0.18}$\pm$\textbf{0.05} & 1.40            & 0.354           \\
                  \bottomrule
            \end{tabular}
      \end{tabular}
      \caption{Prediction results for EEG dataset of different models. Standardized mean squared error (SMSE) and Mean Log Loss (MLL)~\cite{rasmussen2003gaussian} comparison (lower is better). Models marked with \textsuperscript{\textdagger} indicate cited results from~\cite{requeima2019gaussian}.}\label{tb:eeg_results_extended}
\end{table}

\end{document}